%% file: main.tex
\documentclass[runningheads]{llncs}
\usepackage{graphicx}
\usepackage[T5]{fontenc}
\usepackage[utf8]{inputenc}
\usepackage{cite}
\usepackage{amssymb}
\usepackage{amsmath}
\usepackage{url}
\usepackage{tabularx}
\usepackage{pgfplots}
\usepackage{float}
\usepackage{sidecap}
\usepackage{wrapfig}
\usepackage{hyperref}
\usepackage{makecell}
\usepackage{appendix}
\usepackage{longtable}
\usepackage{caption}
\usepackage{subcaption}
\usepackage{booktabs}
\usepackage{multirow}
\usepackage{tablefootnote}

\begin{document}


\title{SA2SL: From Aspect-Based Sentiment Analysis to Social Listening System for Business Intelligence}
\titlerunning{From Sentiment Analysis to Social Listening for Business Intelligence}
%
\author{Luong Luc Phan\inst{1,2,3} \and
Phuc Huynh Pham\inst{1,2,3} \and
Kim Thi-Thanh Nguyen\inst{1,2,3} \and Sieu Khai Huynh\inst{1,2,3} \and Tham Thi Nguyen \inst{1,2,3} \and Luan Thanh Nguyen \inst{1,2,3} \and Tin Van Huynh \inst{1,2,4} \and Kiet Van Nguyen \inst{1,2,4,}\thanks{Corresponding author}}
\institute{University of Information Technology, Ho Chi Minh City, Vietnam\and
Vietnam National University, Ho Chi Minh City, Vietnam
\and  
\email{\{18521073,18521260,18520963,18520348, 18521384,17520721\}@gm.uit.edu.vn} \and\email{\{tinhv, kietnv\}}@uit.edu.vn}
%
\authorrunning{Phan et al.}

\maketitle              
\begin{abstract}
In this paper, we present a process of building a social listening system based on aspect-based sentiment analysis in Vietnamese, from creating a dataset to building a real application. Firstly, we create UIT-ViSFD, a {\bf Vi}etnamese {\bf S}martphone {\bf F}eedback {\bf D}ataset, as a new benchmark dataset built based on a strict annotation scheme for evaluating aspect-based sentiment analysis, consisting of 11,122 human-annotated comments for mobile e-commerce, which is freely available for research purposes. We also present a proposed approach based on the Bi-LSTM architecture with the fastText word embeddings for the Vietnamese aspect-based sentiment task. Our experiments show that our approach achieves the best performances (in F1-score) of 84.48\% for the aspect task and 63.06\% for the sentiment task, which performs several conventional machine learning and deep learning systems. Lastly, we build SA2SL, a social listening system based on the best performance model on our dataset, which will inspire more social listening systems in the future. 


\end{abstract}

\input{Section/1-introduction.tex}

\input{Section/2-relatedwork.tex}

\input{Section/3-dataset.tex}

\input{Section/4-approaches}

\input{Section/5-experiments.tex}

\input{Section/6-sociallistening.tex}

\input{Section/7-conclusion}

\bibliographystyle{splncs04}
\bibliography{main}

\end{document}

%% file: Section/1-introduction.tex
\section{Introduction}
\label{introduction}
Sentiment Analysis (SA) is a significant task and widely applied in many fields such as education, commerce, and marketing. However, a regular SA system may not seem sufficient for business organizations and customers. Simple SA systems consisting of three classes: positive, negative, and neutral, have apparent weaknesses that make them difficult to apply in reality. While enterprises expect an accurate system, the SA systems cannot accurately predict if the sentence does not explicitly express a clear sentiment or an opinion. Aspect-Based Sentiment Analysis (ABSA), an extended research form of SA, has the ability to identify sentiments of specific aspects, features, or entities extracted from user comments or feedback.

According to Statista Research Department\footnote{https://www.statista.com/forecasts/1145936/smartphone-users-in-vietnam\#statisticContainer}, in 2017, the number of smartphone users in Vietnam was estimated to reach approximately 28.77 million. This indicates that around 31 percent of the population used a smartphone at this time, with this share predicted to rise to 40 percent by 2021. In Vietnam, smartphones are used for more than just making and receiving phone calls; they are also used for work, communication, entertainment, and shopping. The phone is chosen differently depending on the needs and purposes of the user. Seeing the potential of the ABSA task on the smartphone data domain in the Vietnamese, we build UIT-ViSFD, a {\bf Vi}etnamese {\bf S}martphone {\bf F}eedback {\bf D}ataset for evaluating ABSA. To ensure the data is plentiful and accurate, we collect feedback from a popular e-commerce website in Vietnam. 

High-quality and large-scale datasets are essential in natural language processing for low-resource languages like Vietnamese. Hence, we aim to build a dataset and implement an approach using machine learning techniques for a social listening system. The task is described as follows: the input of the task is a textual comment related to smartphones that customers generate on social media, outputs are aspects of smartphones and their sentiments are mentioned in the textual comment. Several examples are presented in Table \ref{tab:2}.

In this paper, we have three main contributions summarized below.
\begin{itemize}
    \item We present UIT-ViFSD, a new benchmark Vietnamese dataset for evaluating ABSA in mobile e-commerce, consisting of 11,122 human-annotated comments with two tasks: aspect detection and sentiment classification. Our dataset is freely available for research purposes.
    \item We propose an approach using the Bi-LSTM for the Vietnamese ABSA, achieving the best F1-score performances: 84.48\% for the aspect detection and 63.06\% for the sentiment detection, which performs other systems based on conventional machine learning (Naive Bayes, SVM, and Random Forest) and other deep learning models (LSTM and CNN).
    \item We propose SA2SL, a new social listening system based on ABSA for Vietnamese mobile e-commerce texts, which is the basis for making purchase decisions for users and the evidence for managers to improve their products and services.
\end{itemize}

%% file: Section/2-relatedwork.tex
\section{Related Work}
\label{related_works}

SA is a vibrant field to create many studies and their applications in various fields such as economics, politics, and education. In particular, there are a variety of datasets and methods built in different languages and domains. For English, datasets are  available for electric fields \cite{bhowmick-etal-2008-agreement}; books, equipment kitchen, and electronic products \cite{10.3115/1220575.1220618}. 
Besides, a range of competitions in SemEval 2014 Task 4 \cite{pontiki-etal-2014-semeval}, SemEval 2015 Task 12 \cite{pontiki-etal-2015-semeval}, and SemEval 2016 Task 5 \cite{pontiki-etal-2016-semeval} attracted significant attention.

Although Vietnam has nearly 100 million people, Vietnamese is a low-resource language. The research works in the field of SA in Vietnamese such as student feedback detection \cite {8573337}, hate speech detection \cite {van2019hate}, emotion analysis \cite{ho2019emotion}, constructive and toxic detection \cite {nguyen2021constructive}, and complaint classification \cite {nhungnguyen}. However, these tasks are relatively not as complicated as the ABSA task. The first ABSA shared-task in Vietnamese was organized by the Vietnamese Language and Speech Processing (VLSP) community in 2018 \cite{articlevlsp}. Nguyen et al. \cite{articlevlsp} created datasets for studying two tasks: aspect detection and sentiment classification in the hotel and restaurant. Nguyen et al. \cite{8919448} proposed the dataset on the same domains as restaurant and hotel. 

We aim to create a high-quality dataset about smartphones to evaluate the ABSA task in Vietnamese. The smartphone is a top-rated commercial product that is still thriving today. The amount of feedback data from users about the smartphone is enormous and has great potential for exploitation. Therefore, we review several studies related to this domain. The competition SemEval 2016: Task 5\footnote{https://alt.qcri.org/semeval2016/task5/} introduced a couple of datasets in Chinese and English. Singh et al. \cite{7877474} presented a dataset with many aspects: camera, OS, battery, processor, screen, size, cost, storage aspect, and two sentiments labels: positive, negative for three types of phones: iPhone 6, Moto G3, and Blackberry.  Yiran et al. \cite{10.1145/3340997.3341012} built a dataset with aspects such as display, battery, camera, and three sentiment labels: positive, negative, and neutral. In Vietnamese, Mai et al. \cite{mai2018aspect} proposed a small dataset including 2,098 annotated comments about smartphones at the sentence level, not enough to evaluate current SOTA models. As a result, our dataset is more complex and extensive than the previous Vietnamese dataset \cite{mai2018aspect}. Inspired from previous studies \cite{zhou2016text,do2019hate}, we also propose an approach using Bi-LSTM for Vietnamese aspect-based sentiment analysis. From the best performance of this approach and the study \cite{chaturvedi2017sentiment}, we present a new system based on aspect-based sentiment analysis for business intelligence.

%% file: Section/3-dataset.tex
\section{Dataset Creation}
\label{dataset}

The creation process of our dataset comprises five different phases. First, we collect comments from a well-known e-commerce website for smartphones in Vietnam (see Section \ref{datapreparation}). Secondly, we build annotation guidelines for annotators to determine aspects and their sentiments and how to annotate data correctly (see Section \ref{3.2guidelines}). Annotators are trained with the guidelines and annotate data for two tasks in the two following steps: aspect detection and aspect polarity classification (see Section \ref{annotationprocess}). The inter-annotator agreement (IAA) of annotators in the training process is ensured that it reaches over 80\% before performing data annotation independently. Finally, we provide an in-depth analysis of the dataset that helps AI programmers or experts choose models and features suitable for this dataset (see Section \ref{datasetstatistics}).

\subsection{Data Preparation}
\label{datapreparation}
We crawl textual feedback from customers on a large e-commerce website in Vietnam. To ensure diverse and valued data, we collect feedback from the top ten popular smartphone brands used in Vietnam. There are various long comments, rambling reviews, and contradictory reviews, which are ambiguous to understand to determine the correct label of them. Therefore, the comments that are longer than 250 tokens (makes up a very small rate) are removed. We also delete comments that contain too many misspellings in them, which are not easy to understand and annotate correctly.

\subsection{Annotation Guidelines}
\label{3.2guidelines}
Data annotation is performed by five annotators who follow annotation guidelines and a strict annotating process to ensure data quality. Annotators determine aspects of each comment and then annotate their sentiment polarity labels: positive (Pos), neutral (Neu), and negative (Neg). Table \ref{tab:2} summarizes all aspects (10 aspects) and sentiment polarities (3 sentiment polarities) in the guidelines, including illustrative examples. For some comments that do not relate to any aspect or do not evaluate the product, we annotate an OTHERS label for these cases which do not express the sentiment. 

\begin{table}[]
\caption{The annotation guidelines for labeling the aspects and their sentiment.}
\centering
\label{tab:2}
\resizebox{\columnwidth}{!}{
\begin{tabular}{l|l|l|c}
\hline
\multicolumn{1}{c|}{\textbf{Aspect}}                                              & \multicolumn{1}{c|}{\textbf{Mean}}                                                                                                               & \multicolumn{1}{c|}{\textbf{User comments}}                                                                                       & \multicolumn{1}{c}{\textbf{Sentiment}}                                                                                 \\ \hline
\textbf{SCREEN}                                                                    & \begin{tabular}[c]{@{}l@{}}User comments express screen quality, size, colors, and display technology.\end{tabular}                                   & \begin{tabular}[c]{@{}l@{}}màn hình đẹp\\ (a nice screen)\end{tabular}                       & \begin{tabular}[c]{@{}l@{}}Pos\end{tabular}                                        \\ \hline
\textbf{CAMERA}                                                                    & \begin{tabular}[c]{@{}l@{}}The comments mention the quality of a camera, vibration, delay, \\focus, and image colors.\end{tabular}                   & \begin{tabular}[c]{@{}l@{}}điện thoại chụp hình mờ\\ (the phone took blur picture)\end{tabular}                                 & \begin{tabular}[c]{@{}l@{}}Neg\end{tabular}                           \\ \hline
\textbf{FEATURES}                                                                  & \begin{tabular}[c]{@{}l@{}}The users refer to features, fingerprint sensor,\\  wifi connection, touch and face detection of the phone.\end{tabular} & \begin{tabular}[c]{@{}l@{}}nhận diện khuôn mặt chậm\\ (the face detection is slow)\end{tabular} & \begin{tabular}[c]{@{}l@{}}Neg\end{tabular}                                                  \\ \hline
\textbf{BATTERY}                                                                   & The comment describes battery capacity and battery quality.                                                                                         & \begin{tabular}[c]{@{}l@{}}pin trâu\\ (long battery life)\end{tabular}                                                             & \begin{tabular}[c]{@{}l@{}}Pos\end{tabular}                                                \\ \hline
\textbf{PERFOMANCE}                                                                & \begin{tabular}[c]{@{}l@{}}The reviews describe ramming capacity, processor chip, \\ performance using, and smoothness of the phone.\end{tabular}    & \begin{tabular}[c]{@{}l@{}}cấu hình có thể chấp nhận được\\ (acceptable configuration )\end{tabular}             & \begin{tabular}[c]{@{}l@{}}Neu\end{tabular}                                              \\ \hline
\textbf{STORAGE}                                                                   & \begin{tabular}[c]{@{}l@{}}The comment mention storage capacity, the ability to \\ expand capacity through memory cards.\end{tabular}            & \begin{tabular}[c]{@{}l@{}}bộ nhớ lớn\\ (large storage)\end{tabular}                                                          & \begin{tabular}[c]{@{}l@{}}Pos\end{tabular}                                                  \\ \hline
\textbf{DESIGN}                                                                    & The reviews refer to the style, design, and shell.                                                                                               & \begin{tabular}[c]{@{}l@{}}điện thoại thiết kế thô\\ (rough design phone)\end{tabular}                                      & \begin{tabular}[c]{@{}l@{}}Neg\end{tabular}                                              \\ \hline
\textbf{PRICE}                                                                     & The comments present the specific price of the phone.                                                                                           & \begin{tabular}[c]{@{}l@{}}giá cả ở mức trung bình\\ (the price is at average)\end{tabular}                                                               & \begin{tabular}[c]{@{}l@{}}Neu\end{tabular}                                              \\ \hline
\textbf{GENERAL}                                                                   & The reviews of customers generally comment about the phone.                                                                                            & \begin{tabular}[c]{@{}l@{}}mọi thứ đều ok\\ (everything is ok)\end{tabular}                                        & \begin{tabular}[c]{@{}l@{}}Pos\end{tabular} \\ \hline
\textbf{\begin{tabular}[c]{@{}l@{}}SER\&ACC\tablefootnote{SER\&ACC is short for SERVICE and ACCESSORIES.}\end{tabular}} & \begin{tabular}[c]{@{}l@{}}The comments mention sales service, warranty, and \\ review of accessories of the phone.\end{tabular}                 & \begin{tabular}[c]{@{}l@{}}nhân viên tư vấn nhiệt tình\\ (shop assistants advice enthusiastic)\end{tabular}                             & \begin{tabular}[c]{@{}l@{}}Pos\end{tabular}                                    \\ \hline
\end{tabular}}
\end{table}

\subsection{Annotation Process}
\label{annotationprocess} 

\begin{figure}[H]
\centering
\includegraphics[scale=0.4]{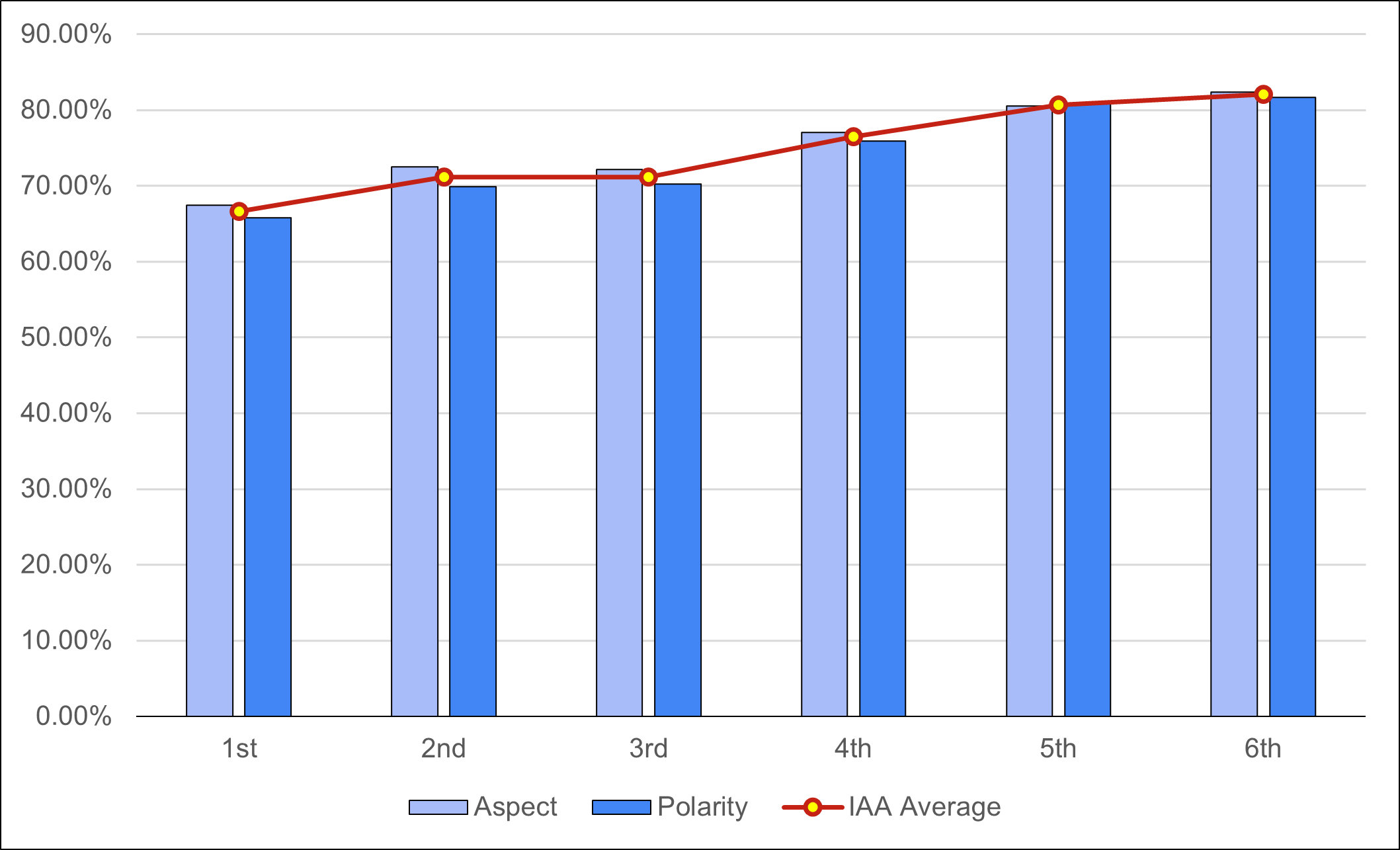}
\caption{\label{fig2}Inter-annotator agreements of six different annotation training rounds.}
\end{figure}

Annotators spend six training rounds to obtain a high inter-annotator agreement, and the strict guidelines are complete. In the first round, after building the first guidelines, our annotators annotate 200 comments together to understand the principles of data annotation. For the five remaining rounds, each round, we randomly take a set of 200 comments and individually annotate these 200 comments. For disagreement cases, we decide the final label by discussing and having a poll among annotators. These causes of disagreement are discussed and corrected in the guidelines; we also add cases that the guidelines have not covered after thorough discussion. The six training rounds resulted in a high inter-annotator agreement of the team and sufficiently completed full guidelines to achieve the dataset. The inter-annotator agreement is estimated by Cohen's Kappa coefficient \cite{bhowmick-etal-2008-agreement}. The formula is described as follows. 
\begin{equation}
k=\frac{Pr(a)-Pr(e)}{1-Pr(e)} 
\end{equation}
Where \emph{k} is the annotator agreement, \emph{Pr(a)} is the relative observed agreement among raters, and \emph{Pr(e)} the hypothetical probability of chance agreement.
To ensure the dataset quality, we calculate inter-annotator agreements of pairs of team members that annotate both the aspect annotation task and the sentiment annotation task. Until the inter-annotator agreement of all labels reaching over 80\% and completing the annotation guidelines, annotators have labeled the comments independently. During this annotating phase, in case of encountering difficult feedback, we discuss together the correct annotation and then revise the guidelines to obtains more high-quality guidelines. Figure \ref{fig2} shows the inter-annotator agreements on two tasks during training phases.

\label{sec:dl}

\subsection{Dataset Statistics}
\label{datasetstatistics}
Our dataset consists of 11,122 comments, including five columns: index (row number), comment (comments), n\_star (customer star ratings), date\_time (comment time), and label (label of a comment). We randomly divide the dataset into three sets: the training (Train), development (Dev), and test (Test) sets in the 7:1:2 ratio.

\begin{table}[H]
\caption{\label{tab:4} Overview statistics of Train/Dev/Test sets of our dataset.}
\centering
\footnotesize
\begin{tabular}{l|r|r|r}
\hline
                                      & \multicolumn{1}{c|}{\textbf{Train}} & \multicolumn{1}{c|}{\textbf{Dev}} & \multicolumn{1}{c}{\textbf{Test}} \\ \hline
\textbf{Number of Comments}           & 7,786                               & 1,112                             & 2,224                              \\ \hline
\textbf{Number of Tokens}             & 283,460                             & 39,023                            & 80,787                             \\ \hline
\textbf{Number of Aspects}            & 23,597                              & 3,371                             & 6,742                              \\ \hline
\textbf{Average number of aspects per sentence} & 3.3                                & 3.2                              & 3.3                               \\ \hline
\textbf{Average length per sentence}  & 36.4                               & 35.1                             & 36.3                              \\ \hline
\end{tabular}
\end{table}


Table \ref{tab:4} presents overview statistics of our dataset. The splitting ratio corresponds to the number of words and labels in the Train, Dev, and Test sets. Each comment has three aspect labels and is approximately 36 tokens on average.
\begin{table}[H]
\caption{The distribution of aspects and their sentiments of our dataset.}\label{tab:5}
\resizebox{\columnwidth}{!}{
\begin{tabular}{l|r|r|r|r|r|r|r|r|r|r}
\hline
\multicolumn{1}{c|}{\multirow{2}{*}{\textbf{Aspect}}} & \multicolumn{3}{c|}{\textbf{Train}}                                                                                     & \multicolumn{3}{c|}{\textbf{Dev}}                                                                                       & \multicolumn{3}{c|}{\textbf{Test}}                                                                                      & \multicolumn{1}{c}{\multirow{2}{*}{\textbf{Total}}} \\ \cline{2-10}
\multicolumn{1}{c|}{}                                 & \multicolumn{1}{c|}{\textbf{Pos}} & \multicolumn{1}{c|}{\textbf{Neu}} & \multicolumn{1}{c|}{\textbf{Neg}} & \multicolumn{1}{c|}{\textbf{Pos}} & \multicolumn{1}{c|}{\textbf{Neu}} & \multicolumn{1}{c|}{\textbf{Neg}} & \multicolumn{1}{c|}{\textbf{Pos}} & \multicolumn{1}{c|}{\textbf{Neu}} & \multicolumn{1}{c|}{\textbf{Neg}} & \multicolumn{1}{c}{}                                \\ \hline
\textbf{BATTERY}                                       & 2,027                                  & 349                                   & 1,228                                  & 303                                    & 51                                    & 150                                    & 554                                    & 92                                    & 368                                    & 5,122                                                \\ \hline
\textbf{CAMERA}                                        & 1,231                                  & 288                                   & 627                                    & 172                                    & 36                                    & 88                                     & 346                                    & 71                                    & 171                                    & 3,030                                                \\ \hline
\textbf{DESIGN}                                        & 999                                    & 77                                    & 302                                    & 135                                    & 12                                    & 40                                     & 274                                    & 28                                    & 96                                     & 1,963                                                \\ \hline
\textbf{FEATURES}                                      & 785                                    & 198                                   & 1,659                                  & 115                                    & 33                                    & 233                                    & 200                                    & 52                                    & 459                                    & 3,734                                                \\ \hline
\textbf{GENERAL}                                       & 3,627                                  & 290                                   & 949                                    & 528                                    & 34                                    & 127                                    & 1,004                                  & 83                                    & 294                                    & 6,936                                                \\ \hline
\textbf{PERFORMANCE}                                   & 2,253                                  & 391                                   & 1,496                                  & 327                                    & 45                                    & 210                                    & 602                                    & 116                                   & 454                                    & 5,894                                                \\ \hline
\textbf{PRICE}                                         & 609                                    & 391                                   & 316                                    & 72                                     & 144                                   & 36                                     & 162                                    & 328                                   & 79                                     & 2,882                                                \\ \hline
\textbf{SCREEN}                                        & 514                                    & 56                                    & 379                                    & 62                                     & 12                                    & 47                                     & 136                                    & 17                                    & 116                                    & 1,339                                                \\ \hline
\textbf{SER\&ACC}                                      & 1,401                                  & 107                                   & 487                                    & 199                                    & 13                                    & 78                                     & 199                                    & 27                                    & 167                                    & 2,678                                                \\ \hline
\textbf{STORAGE}                                       & 59                                     & 107                                   & 21                                     & 11                                     & 1                                     & 2                                      & 18                                     & 3                                     & 6                                      & 132                                                  \\ \hline
\textbf{Total}                                         & 13,505                                 & 2,903                                 & 7,464                                  & 1,924                                  & 381                                   & 1,011                                  & 3,495                                  & 817                                   & 2,210                                  &                                                      \\ \hline
\end{tabular}
}
\end{table}

Table \ref{tab:5} describes the distribution of aspects and their sentiment in the Train, Dev, and Test sets of our dataset. Through our analysis, the dataset has an uneven distribution on both the aspect and sentiment labels. While some aspect labels have many data points, another has a negligible number (the General aspect has 6,936 data points compared to the Storage with 132 annotated comments). In addition, we notice a significant difference between the three sentiment polarities. Positive accounts for the most significant number of 56.13\% of the total number of labels followed the negative polarity with 31.70\%, whereas the neutral polarity only accounts for 12.17\%. Our dataset is inbalanced and includes diverse comments on different smartphone products on social media, which is challenging to evaluate ML algorithms on social media texts.

%% file: Section/4-approaches.tex
\section{Our Approach}

\begin{figure}[H]
\centering
\includegraphics[scale=0.37]{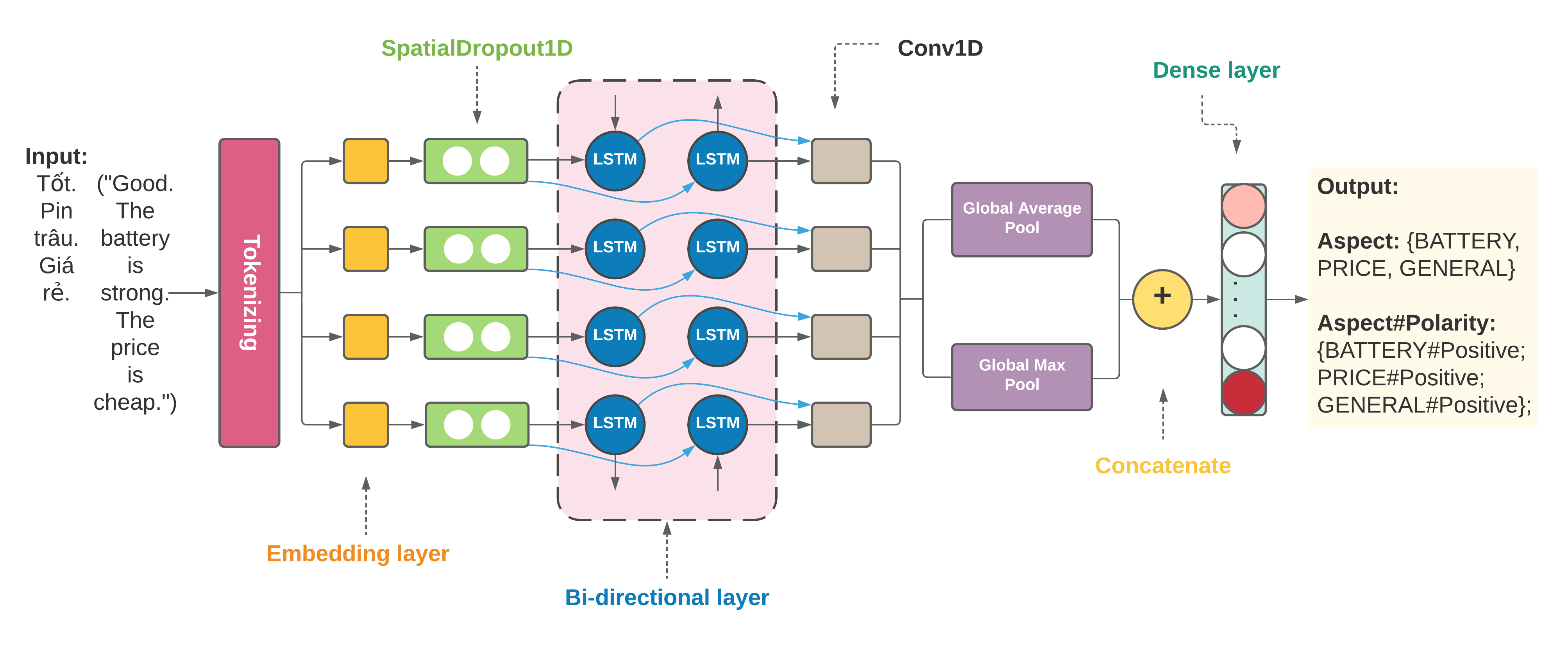}
\caption{An overview of our Vietnamese ABSA system using Bi-LSTM.}\label{fig4}
\end{figure}
Inspired by Bi-LSTM for text classification \cite{zhou2016text}, we propose an approach using the Bi-LSTM model for Vietnamese ABSA. The overview architecture is depicted in Figure \ref{fig4}. This architecture consists of a tokenizer, embedding layer, SpatialDropout1D layer, Bi-LSTM layer, convolutional layer, two pooling layers, and a dense layer. First, the comment goes through a tokenizer, which converts each 
token in the comment to an integer value based on the vocabulary index. Then, they are processed through an embedding layer to convert into representative vectors. The architecture uses the fastText embeddings \cite{DBLP:journals/corr/BojanowskiGJM16} as input token representations. fastText has good token representations, and it encodes for rare tokens that do not appear during training and is a good selection for Vietnamese social media texts \cite{huynh2020simple}. To minimize overfitting, we utilize SpatialDropout1D to lower the parameters after each train. We employ a Bi-LSTM layer to extract abstract features, which is made up of two LSTMs with their outputs stacked together. The comment is read forward by one LSTM and backward by the other. We concatenate the hidden states of each LSTM after they have processed their respective final token. The Bi-LSTM uses two separate LSTM units, one for forward and one for backward. A convolutional layer is used to convert a multi-dimensional matrix from Bi-LSTM to a 1-dimensional matrix. The pooling layer consists of two parallel layers: the global average pool and the global max pool. The function of the pooling layer is conducted to reduce the spatial size of the representation. The idea is to choose the highest element and the average element of the feature map to extract the most salient features from the convolutional layer. Finally, the labels of the two sub-tasks for the ABSA task is obtained after normalizing in the dense layer.

%% file: Section/5-experiments.tex
\section{Experiments}
\label{experiment}

\subsection{Baseline Systems}

We compare the proposed approach with the following baselines. For traditional machine learning, we experiment with a system based on Naive Bayes \cite{articlenaive}, Support Vector Machine (SVM) \cite{ALSMADI2018386}, and Random Forest \cite{Breiman2001}, which are popular methods for text classification. For deep neural networks, we implement systems based on Convolutional Neural Networks (CNN)  \cite{dos-santos-gatti-2014-deep} and Long Short-Term Memory (LSTM) \cite{ruder-etal-2016-hierarchical}, which achieved SOTA results in different NLP tasks.

\subsection{Experimental Results}
\label{result}

There are two phases to measure the ABSA systems: aspect detection and sentiment prediction. We use the precision, recall, and F1-score (macro average) to measure the performance of models. We do not detect sentiments of OTHERS aspect because they cannot show their sentiments, so we give it the NaN value. 

Table \ref{tab:6} presents overall results on aspect-based and sentiment-based tasks on machine learning systems. According to our results, we can see that deep learning models have significantly better performance than traditional machine learning models. In particular, Bi-LSTM achieves the best F1-scores of 84.48\% and 63.06\% for the aspect detection and sentiment prediction, whereas SVM shows the lowest performances. Besides, the sentiment detection task makes it difficult for the first models when the F1-score are low (below 70\%). There is a considerable discrepancy between deep learning and machine learning models. In particular, the deep learning model (Bi-LSTM) obtains the best F1-score of 84.48\%, whereas the best machine learning model (Naive Bayes) only gains 64.65\% in F1-score. 

\begin{table}[H]
\caption{Performances (\%) of different ABSA systems.}\label{tab:6}
\centering
\begin{tabular}{l|r|r|r|r|r|r}
\hline
\multicolumn{1}{c|}{\multirow{2}{*}{\textbf{System}}} & \multicolumn{3}{c|}{\textbf{Aspect Detection}}                                                                          & \multicolumn{3}{c}{\textbf{Sentiment Detection}}                                                                       \\ \cline{2-7} 
\multicolumn{1}{c|}{}                                 & \multicolumn{1}{c|}{\textbf{Precision}} & \multicolumn{1}{c|}{\textbf{Recall}} & \multicolumn{1}{c|}{\textbf{F1-score}} & \multicolumn{1}{c|}{\textbf{Precision}} & \multicolumn{1}{c|}{\textbf{Recall}} & \multicolumn{1}{c}{\textbf{F1-score}} \\ \hline
\textbf{Naive Bayes}                                   & 72.18                                   & 59.53                                & 64.65                                  & 49.07                                   & 30.43                                & 37.56                                  \\ 
\textbf{SVM}                                           & 36.45                                   & 51.34                                & 42.63                                  & 16.09                                   & 23.35                                & 19.69                                  \\ 
\textbf{Random Forest}                                 & 45.72                                   & 50.15                                & 47.83                                  & 17.11                                   & 24.56                                & 20.17                                  \\ \hline
\textbf{CNN}                                           & 77.25                                   & 63.49                                & 69.70                                  & 33.34                                   & 22.92                                & 27.16                                  \\ 
\textbf{LSTM}                                          & 82.61                                   & 78.05                                & 80.27                                  & 56.51                                   & 48.39                                & 52.13                                  \\ \hline
\textbf{Our Approach}                                       & 87.55                                   & 83.22                                & {\bf 84.48}                                  & 65.82                                   & 60.53                                & {\bf 63.06}                                  \\ \hline
\end{tabular}
\end{table}

The results of the Bi-LSTM on the aspects and sentiment classification are shown in Table \ref{tab:7}. 
As mentioned above, the sentiment detection performance of the Bi-LSTM is lower than that of the aspect detection results (F1-score of 84.48\% for the aspect detection task and that of 63.06\% for the sentiment detection task). In the aspect detection task, the Bi-LSTM system also achieves relatively high and positive results (F1-score for all aspects is 60\% higher, and many aspects have F1-score below 80\%). On the contrary, the system performance on the sentiment-aspect detection task is relatively low (F1-score for all aspects is below 75\% and the aspect with the highest F1-score is Camera with 74.69\%). In terms of aspects detection, the highest is the Battery with 95.00\% F1-score. As for the sentiment detection, the Storage aspect polarity is only 30.10\% F1-score. The Storage label result explains the lack of quantity uniformity in the labels (the Storage aspect only covers 1.19\% of the dataset). These results are pretty interesting to explore further models on this dataset. In general, the Bi-LSTM system outperforms other algorithms when it comes to detecting aspects and their sentiments. However, their ability to extract sentiment features for each aspect is limited in all machine learning models, which will be exploited in future work.

\begin{table}[H]
\centering
\caption{Performances (\%) of our approach in terms of different aspects.}\label{tab:7}
\begin{tabular}{l|r|r|r|r|r|r}
\hline
\multicolumn{1}{c|}{\multirow{2}{*}{\textbf{Aspect}}} & \multicolumn{3}{c|}{\textbf{Aspect Detection}}                                                                          & \multicolumn{3}{c}{\textbf{Sentiment Detection}}                                                                       \\ \cline{2-7} 
\multicolumn{1}{c|}{}                                 & \multicolumn{1}{c|}{\textbf{Precision}} & \multicolumn{1}{c|}{\textbf{Recall}} & \multicolumn{1}{c|}{\textbf{F1-score}} & \multicolumn{1}{c|}{\textbf{Precision}} & \multicolumn{1}{c|}{\textbf{Recall}} & \multicolumn{1}{c}{\textbf{F1-score}} \\ \hline
\textbf{Screen}                                        & 89.41                                   & 85.22                                & 87.26                                  & 64.22                                   & 60.50                                & 62.30                                  \\ \hline
\textbf{Camera}                                        & 85.35                                   & 85.00                                & 85.17                                  & 76.22                                   & 73.23                                & \textbf{74.69}                         \\ \hline
\textbf{Features}                                      & 89.01                                   & 88.29                                & 88.64                                  & 70.10                                   & 60.20                                & 64.77                                  \\ \hline
\textbf{Battery}                                       & 95.00                                   & 94.21                                & \textbf{94.60}                         & 73.33                                   & 72.14                                & 72.73                                  \\ \hline
\textbf{Performance}                                   & 89.22                                   & 88.33                                & 88.77                                  & 68.44                                   & 65.15                                & 66.75                                  \\ \hline
\textbf{Storage}                                       & 84.11                                   & 70.12                                & 76.48                                  & 32.32                                   & 28.18                                & 30.10                                  \\ \hline
\textbf{Design}                                        & 89.30                                   & 86.46                                & 87.85                                  & 72.35                                   & 70.71                                & 71.41                                  \\ \hline
\textbf{Price}                                         & 90.28                                   & 90.12                                & 90.19                                  & 72.13                                   & 70.71                                & 71.41                                  \\ \hline
\textbf{General}                                       & 82.28                                   & 81.17                                & 81.72                                  & 67.18                                   & 64.34                                & 65.72                                  \\ \hline
\textbf{Ser\&Acc}                                      & 88.19                                   & 86.16                                & 87.16                                  & 62.23                                   & 56.50                                & 59.22                                  \\ \hline
\textbf{Others}                                        & 61.34                                   & 60.31                                & 60.82                                  & NaN                                     & NaN                                  & NaN                                    \\ \hline
\textbf{Macro Avg}                                     & 85.77                                   & 83.22                                & \textbf{84.48}                         & 65.82                                   & 60.53                                & \textbf{63.06}                         \\ \hline
\end{tabular}
\end{table}

%% file: Section/6-sociallistening.tex
\section{SA2SL: Social Listening System Using ABSA}
\label{application}

\begin{figure}[H]
    \centering
    \includegraphics[scale=0.25]{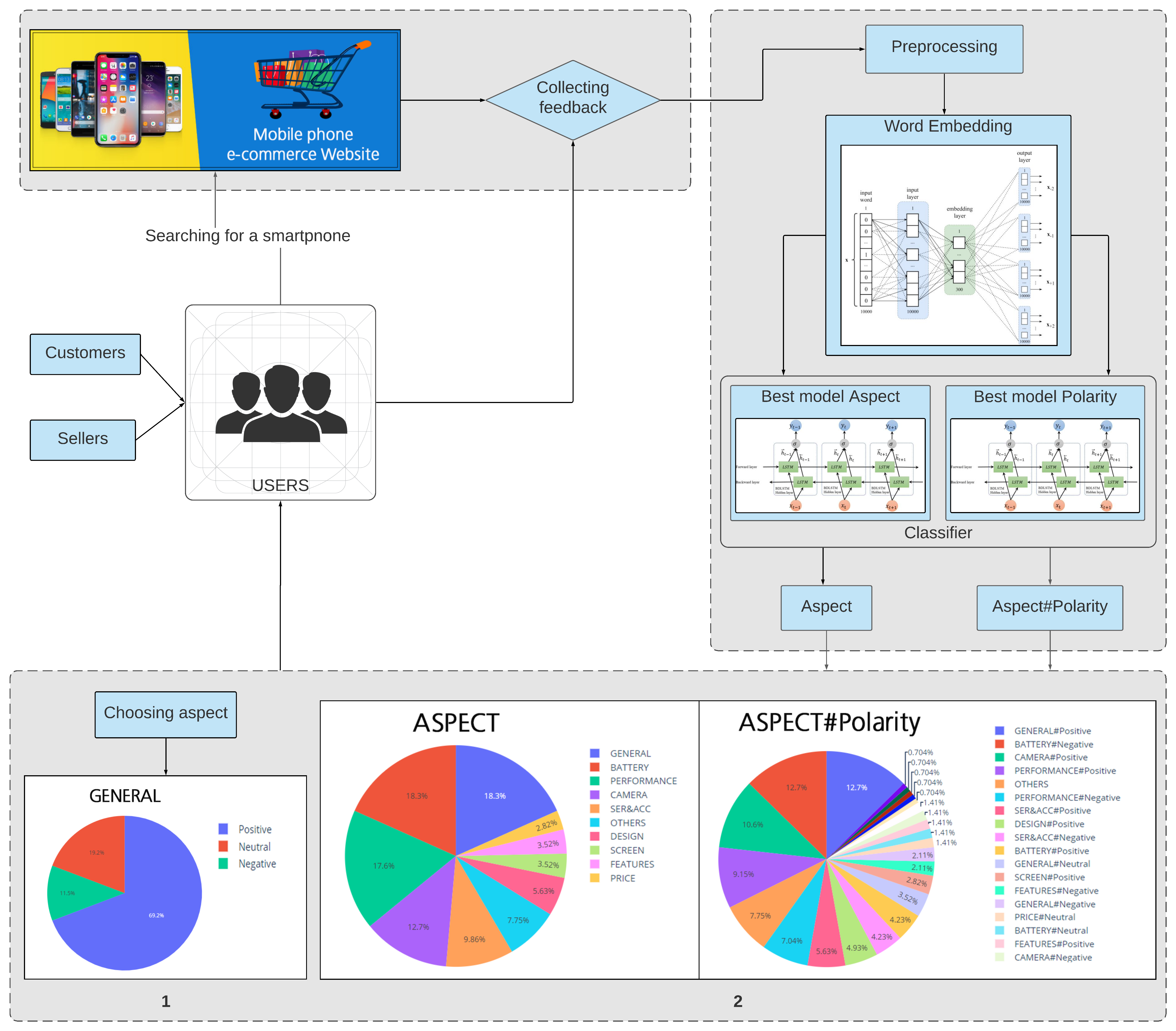}
    \caption{Social listening architecture using ABSA for smartphone products.}\label{fig3}
\end{figure}

Inspired by the best performance results and the investigation \cite{chaturvedi2017sentiment}, we propose SA2SL, a social listening system architecture based on Vietnamese ABSA for analyzing what customers discuss about products on social media. Figure \ref{fig3} depicts a social listening system for smartphones that uses aspect-based sentiment analysis. This application assists customers and business companies in automatically categorizing comments and determining consumer perspectives. The application assists shoppers in selecting a phone that meets their specific requirements. Moreover, manufacturers can focus on the needs, expectations of the customers and propose suitable improvement options to improve product quality in future. Aspect sentiment analysis is crucial because it can assist businesses in automatically sorting and analyzing consumer data, automating procedures such as customer service activities, and gaining valuable insights. 

Firstly, the user selects the name of the phone, and the application collects all comments on that phone. Secondly, we do the pre-processing of the comments, and then we feed them into word embedding, and they become vectors. Next, the vectors are then analyzed using two models: aspect detection and sentiment detection (aspect\#sentiment). The input is a list of comments, and the output is aspects and their sentiments. The final analyses are visualized as follows: (1) Depending on which aspect they are interested in, the consumer or business company recognizes the analysis of user feedback regarding the sentiment polarity in the first chart. To clearly understand an aspect, the user of the system can select one of ten aspects, and then the system detail displays the distribution of its sentiment polarities. (2) The second chart describes the proportion (\%) of the predicted aspects and summarizes their sentiments of all comments.

%% file: Section/7-conclusion.tex
\section{Conclusion and Future Work}
\label{conclusion}

We have three main contributions, which are (1) creating UIT-ViSFD, a benchmark dataset of smartphone feedback for the ABSA task, (2) proposing the approach for this task, and (3) building a social listening system based on the primary technology of ABSA. We experimented with two different types of models: traditional machine learning and deep learning on our dataset to compare with our approach. Our approach outperformed the others, achieving F1-scores of 84.48\% and 63.06\% for aspect detection and sentiment detection, respectively. Although the aspect detection results were positive, the sentiment detection results were relatively low, which is challenging for further machine learning-based systems. Finally, we presented a novel social listening system based on ABSA for the low-resource language like Vietnamese.

NLP experts can develop a new span detection dataset using our ABSA dataset. Besides, we conduct experiments based on powerful aspect-based systems using BERTology models \cite{rogers2020primer}, transfer learning approaches \cite{ruder2019transfer}, and reinforcement learning \cite{gai2018reinforcement}. We utilize the relationship between the rank of input data and the performance of a random weight neural network \cite{cao2020study} to enhance the overall performance of the task. Lastly, we recommend that a social listening system should be integrated with different social media tasks \cite{ho2019emotion,nguyen2021constructive,nhungnguyen} into the social listening system, which benefits social business intelligence.